\documentclass[runningheads]{llncs}
\usepackage{graphicx}
\usepackage{tabularx}
\begin{document}
\title{Measuring Domain Portability and Error Propagation in Biomedical QA}
%
%
\author{Stefan Hosein \and
Daniel Andor \and
Ryan McDonald}
%
%
\institute{Google\\
\email{\{smhosein,andor,ryanmcd\}@google.com}}
\maketitle              
\begin{abstract}
In this work we present Google's submission to the BioASQ 7 biomedical question answering (QA) task (specifically Task 7b, Phase B). The core of our systems are based on BERT QA models, specifically the model of \cite{alberti2019bert}. In this report, and via our submissions, we aimed to investigate two research questions. We start by studying how domain portable are QA systems that have been pre-trained and fine-tuned on general texts, e.g., Wikipedia. We measure this via two submissions. The first is a non-adapted model that uses a public pre-trained BERT model and is fine-tuned on the Natural Questions data set \cite{kwiatkowski2019natural}. The second system takes this non-adapted model and fine-tunes it with the BioASQ training data. Next, we study the impact of error propagation in end-to-end retrieval and QA systems. Again we test this via two submissions. The first uses human annotated relevant documents and snippets as input to the model and the second predicted documents and snippets. Our main findings are that domain specific fine-tuning can benefit Biomedical QA. However, the biggest quality bottleneck is at the retrieval stage, where we see large drops in metrics -- over 10pts absolute -- when using non gold inputs to the QA model.

\end{abstract}

\section{Introduction}
BioASQ \cite{tsatsaronis2015overview} is a large-scale online biomedical research competition. There are many tasks within the competition: question answering (QA), information retrieval and semantic indexing. Our submissions focus on Task 7b, Phase B which requires participating systems to generate ideal or exact answers to biomedical questions using mainly PubMed articles. We focus on exact answers which can include factoid, list, and yes/no question types.

The systems we used for QA were all BERT-based \cite{devlin2018bert} models using the public available \emph{large} pre-trained models and fine-tuned on the Natural Questions corpus \cite{alberti2019bert,kwiatkowski2019natural} and Conversational Question Answering dataset \cite{Reddy2018CoQAAC}. Additionally, three of the four systems we submitted were further fine-tuned on the BioASQ training data. The difference between the biomedical specific models is the input into the models: using only snippets, using snippets from the previous information retrieval phase (Task 7b, Phase A) and a mixture of snippets and abstracts. This work-flow has no pre-processing of the data necessary and uses very little in-domain knowledge to achieve successful results. 

Our systems focused mainly on factoid questions and their results. The evaluation metrics for factoid were strict accuracy, lenient accuracy, and Mean Reciprocal Rank (MRR) \cite{tsatsaronis2015overview}. The results of the competition show that all our models are always in the top half of systems for factoid questions which indicate that neural QA models based on large pre-trained language models are very robust across domains. In addition, since our system used snippets from the previous information retrieval phase and had a lower but still competitive accuracy indicated that the limiting factor of this neural model is the document and snippet retrieval architecture and not the QA model itself.

In this paper we start with a literature review which explains our reasoning for using BERT-based models and the architectures of previous entrants for the BioASQ challenge, then we go in-depth into explaining the differences between our 4 systems that were submitted, lastly we discuss the performance of our systems and how error propagates between retrieval and QA systems.

\section{Related Work}
The use of BERT-based models \cite{devlin2018bert} is becoming ubiquitous in the field of question answering (QA). At the time of this writing, out of the top 5 systems in SQuAD 2.0 \cite{Rajpurkar2016SQuAD10}, 4 are BERT models. For the CoQA \cite{Reddy2018CoQAAC} challenge, all of the top 5 systems are BERT models. With the success of BERT models, many papers are tuning these models to their specific domain. One such paper is BioBERT \cite{lee2019biobert}, where the authors created a  domain specific language representation biomedical BERT model for a few biomedical tasks, one being question answering. They evaluated their models on BioASQ test sets for BioASQ 4, 5 and 6. They saw a an absolute improvement of 9.61\% with the models.

The BioASQ \cite{Tsatsaronis2015AnOO} competition has been very popular amongst researchers. Some of the early systems in BioASQ were not neural architectures. For the 2nd BioASQ challenge, \cite{Papanikolaou2014EnsembleAF} developed a system that tries to extract the lexical answer type of the question. Then, they selected the relevant snippets for each question and provided these as inputs to MetaMap\footnote{https://metamap.nlm.nih.gov/} which extracted candidate answers for each factoid question. For the 3rd iteration of the challenge \cite{Yang2015LearningTA} used a three layer architecture for factoid and list questions. The architecture is based on the framework \cite{Yang2013BuildingOI} and including many components like MetaMap and ClearNLP\footnote{\url{https://github.com/clir/clearnlp/}}. In BioASQ 4 both \cite{Papanikolaou2014EnsembleAF} and \cite{Yang2013BuildingOI} imporoved their models using more biomdeical information into their systems. Neural architecture systems started to appear more frequently from BioASQ 5, with the \textit{DeepQA} systems using the then state-of-the-art QA model, FastQA \cite{Weissenborn2017FastQAAS}. The FastQA was extended by using biomedical word embeddings and pre-training on QA datasets (SQuAD) then fine-tuning on the BioASQ training set. In the last BioASQ challenge (BioASQ 6), there were numerous systems that used neural architectures like LSTMs \cite{Hochreiter1997LongSM,nentidisetal2018results}.

\section{BERT Model}

Recent work on learning word representations have focused on learning context dependent representations. An example, the word \texttt{bank}, it could mean the land alongside the river/lake or a financial establishment. Previous methods would have a single representation of the word \texttt{bank} unlike more modern methods which will have two representations for the word based on its context in the sentence. 
BERT \cite{devlin2018bert} is one such method to produce contextualized word embeddings. The most common instantiaion of BERT is pre-trained using bidirectional transformers to predict randomly masked words in a sequence, thus removing the limitation that previous bidirectional language models had: the fact that future words should not be seen. In addition, BERT predicts the next sentence given a previous sentence and these two tasks allow BERT to obtain state-of-the-art performance on many NLP tasks.

Our QA model follows the Natural Questions (NQ) baseline model \cite{alberti2019bert}, an extractive QA model based on BERT \cite{devlin2018bert}.
In the context of the BioASQ data: given a pair of question (the body) $Q$ and context/body (the snippets or some augmentation of the snippets) $S$, the model predicts the answer by scoring all the sub-spans (candidate answers taken from $S$) and then ranking all these sub-spans by their score. For more in-depth details, see \cite{alberti2019bert}.

\section{Systems Overview}
There were four systems that we submitted for evaluation in BioASQ Task 7b, Phase B. Below is a brief overview of each system, we give more details in further sub-sections.

\begin{figure}[t]
\centering
\includegraphics[scale=0.28]{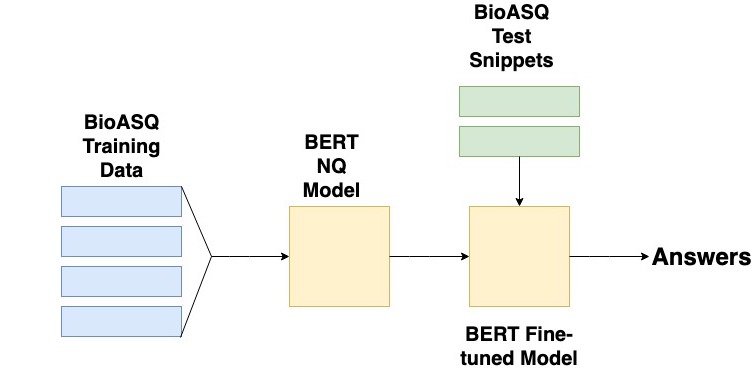}
\caption{A figure showing a model which was fine-tuned on BioASQ training data using provided BioASQ test (gold) snippets from test batches to generate the answers for questions (google-gold-input).}
\label{fig:sa}
\end{figure}

\begin{itemize}
    \item \textbf{google-gold-input}: fine-tuned on BioASQ training data, used the provided gold snippets as input to the QA model (see Figure \ref{fig:sa})
    \item \textbf{google-gold-input-ab}: fine-tuned on BioASQ training data, used the provided gold snippets and the abstract of the top ranked document as input to the QA model
    \item \textbf{google-gold-input-nq}: no in-domain training, used the provided gold snippets as input to the QA model
    \item \textbf{google-pred-input}: fine-tuned on BioASQ training data, used snippets from the top-ranked submission from Task 7b, Phase A as input to the QA model
\end{itemize}

\subsection{No In-Domain Training}
\label{sec:no-in-domain-training}

To give our baseline system, google-gold-input-nq, exposure to a broad set of domains, we trained on both the NQ \cite{kwiatkowski2019natural} and CoQA \cite{Reddy2018CoQAAC} datasets.
Both NQ and CoQA contain Wikipedia data, while CoQA adds four additional domains, covering news and fiction.

After training on NQ as in \cite{alberti2019bert}, we further fine-tuned on CoQA with a learning rate of 5e-5, batch size of 32, for 2 epochs.

\subsection{BioASQ Fine-Tuning}
Two of our models -- accounting for three of our systems -- were fine-tuned using the BioASQ training data. The difference between these two models is that one uses a concatenation of relevant snippets as model context (google-gold-input) while the other uses the abstract of the most relevant document concatenated with any remaining snippets (google-gold-input-ab), see Table~\ref{tab:pretrain} for an example. We used only one abstract as using abstracts from lower ranked documents would dramatically increase the noise-to-signal ratio.

\begin{table}[t]
\centering
\caption{Table showing the differences between the input at test time for each model.}
\label{tab:pretrain}
\begin{tabularx}{\textwidth}{l X}
\hline
\textbf{System} &  \textbf{Context} \\
\hline
google-gold-input &  Vaspin expression is increased in white adipose tissue \textbackslash{}n Visceral adipose tissue-derived serine protease inhibitor (Vaspin) is an adipocytokine that has been shown to exert anti-inflammatory effects and inhibits apoptosis under diabetic conditions.\\
\hline
google-gold-input-ab & Vaspin suppresses cytokine-induced inflammation in 3T3-L1 adipocytes via inhibition of NF\textbackslash{}u03baB pathway.\textbackslash{}n Vaspin expression is increased in white adipose tissue (WAT) of diet-induced obese mice and rats and is supposed to compensate HFD-induced inflammatory processes and insulin resistance in adipose tissue by \textbf{\ldots} \textbackslash{}n Visceral adipose tissue-derived serine protease inhibitor (Vaspin) is an adipocytokine that has been shown to exert anti-inflammatory effects and inhibits apoptosis under diabetic conditions.\\
\hline
\end{tabularx}
\end{table}

Starting with the model trained in Section \ref{sec:no-in-domain-training}, we fine-tuned on the BioASQ training set using a learning rate of 1e-7, batch size of 32, for 10 epochs.
The large number of epochs was necessary due to the very small training dataset size of $\sim 2700$ questions.

\subsection{Snippet Retrieval}
The model, google-gold-input, and the model used for snippet retrieval, google-pred-input, is the same, however, the difference between them is at test time. Instead of using the gold-standard test snippets provided by BioASQ, google-pred-input used snippets from the top ranking submission to Task 7b, Phase A \cite{pappas2019aueb}. This allows us to analyze the effect of information retrieval on the QA system since the only difference between google-pred-input and google-gold-input is the context given to the QA system. 
One interesting property is that the predicted set of snippets is often much larger than the gold set. This is partly due to the nature of the data, where the annotators were tasked with finding enough relevant snippets to support the correct answer -- not all the relevant snippets.


\subsection{Yes/No and List Question Types}
Even though our systems participated in some yes/no and list batches, these were heuristic based and not a core part of our model. For yes/no questions, if \textit{yes} or \textit{no} was present in the candidate answers, then we selected the one with the higher log probability. If we could not find \textit{yes} or \textit{no} in the candidate set, we selected yes by default. For list type questions, we selected the top 5 candidates and split the results into single words or phrases by punctuation and then selected the top 5 results from those. Since these were heuristic based, we do not discuss these results in the paper.

\section{Results}

We took part in the last three batches of Task 7b, Phase B. More specifically: the answers of google-gold-input and google-pred-input were evaluated on batches 3, 4 and 5 and google-gold-input-nq and google-gold-input-ab were evaluated on batches 4 and 5. 
For batch 3 our google-gold-input was always in the top two system scores for all factoid evaluations, while google-pred-input had the lowest place of 6th for factoid evaluations. For batches 4 and 5 our scores were generally in the top ten for factoids. 

For a comparison of the best system's score and our models see Table \ref{tab:comp}. The table alludes to a number of interesting results some we discuss in later subsections. One of those results is that adding abstracts was not significantly helpful and indicates that there is a noise-to-signal issue where the system might get diminishing or negative gains after a certain amount of data is used for the context.

\begin{table}[t]
    \caption{Performance on BioASQ Task 7b, Phase B for batches 3 to 5. The underlined system is the best scoring system from
Google’s entries and bold indicates the system is the top from all official entries in that batch. Best Score is the top scoring entry that is not among Google’s submissions.}
    \label{tab:comp}
    \centering
     \begin{tabular*}{\textwidth}{@{\extracolsep{\fill}} c |ccc |ccc |ccc}
    \multicolumn{1}{c}{} & \multicolumn{3}{c}{Batch 3} & \multicolumn{3}{c}{Batch 4} & \multicolumn{3}{c}{Batch 5}\\
    \multicolumn{1}{c}{} & Strict & Lenient & \multicolumn{1}{c}{MRR} & Strict & Lenient & \multicolumn{1}{c}{MRR} & Strict & Lenient &  \multicolumn{1}{c}{MRR} \\ \hline
    Best Score & 0.4483 & 0.6552 & 0.5115 & 0.5882 & 0.8235 & 0.6912 & 0.2857 &0.5143 &	0.3638 \\
    \hline
    gold-input & \underline{0.4138} & \textbf{0.6552} & \underline{0.5023} & \underline{0.4706} & \underline{0.7059} & \underline{0.5495} & \textbf{0.2857} & \underline{0.3714} & \underline{0.3167} \\
    pred-input & 0.3448 & 0.5517 & 0.4322 & 0.3529 & 0.5882 & 0.4338 & 0.1429 & 0.2857 & 0.2057 \\
    gold-input-ab & - & - & - & \underline{0.4706}	& 0.6471 & 0.5255 & 0.2286 & 0.2857 & 0.2571  \\
    gold-input-nq & - & - & - & \underline{0.4706}	& 0.5882 &  0.5132 & \textbf{0.2857} & \underline{0.3714} & 0.3057  \\\hline
    \end{tabular*}
    \vspace{0.1in}
    
\end{table}

It should be noted that these results are preliminary. Humans have yet to judge the outputs off all participating systems. As a precursor to participating in BioASQ7, we investigated the performance of our model on prior year's data. The advantage of doing this is that the test annotations are much more complete, since they also include all the correct answers from the systems that participated that year.  We compare to two baselines. The first is the the best system that participated in that specific year's challenge. The second is a recent state-of-the-art model BioBERT \cite{lee2019biobert}\footnote{The authors of this system also participated in BioASQ7 and preliminary have the highest scoring submission.}. This model is similar in nature to our model, with some differences. First, it is pre-trained on biomedical data. Second, it is only fine-tuned on the BioASQ training data and does not use any additional fine-tuning data, i.e., natural questions.
Note that all models are comparable: 1) they are trained with the specific training data for the year being tested; and 2) they use provide gold snippets as input.

Table~\ref{tab:prev_comp} shows the results. We can see here that our model is very competitive with previous models on this data, including other BERT-based models. The main take-away here is that adding domain general fine-tuning data (i.e., the Natural Questions data) can lead to gains in performance.

\begin{table}[t]
    \caption{Performance on BioASQ Task 4 and 5b, Phase B averaged over all batches. The bold system is the top scoring.  \emph{Best Participant} and \emph{BioBERT} results are from \cite{lee2019biobert}.}
    \label{tab:prev_comp}
    \centering
     \begin{tabular*}{\textwidth}{@{\extracolsep{\fill}} c |ccc |ccc}
    \multicolumn{1}{c}{} & \multicolumn{3}{c}{BioASQ 4} & \multicolumn{3}{c}{BioASQ 5} \\
    \multicolumn{1}{c}{} & Strict & Lenient & \multicolumn{1}{c}{MRR} & Strict & Lenient & \multicolumn{1}{c}{MRR} \\ \hline
    Best Participant &  0.206 & 0.294 & 0.240 & 0.418 & 0.574 & 0.477\\
    BioBERT \cite{lee2019biobert} & \textbf{0.365} & 0.489 & \textbf{0.411} & 0.416 & 0.540 & 0.463 \\
    \hline
    google-gold-input &  0.311 & \textbf{0.540} & 0.400 & \textbf{0.458} & \textbf{0.615} & \textbf{0.520} \\\hline
    \end{tabular*}
    \vspace{0.1in}
    
\end{table}

\subsection{Domain Portability}

To measure domain portability we investigate the model fine-tuned only on the NQ dataset (google-gold-input-nq) and the model that was further fine-tuned on BioASQ training data (google-gold-input-ab). For this experiment, these models use the top-ranked abstract concatenated with snippets from other documents as input. Results for factoid QA are shown in Table~\ref{tab:domain_port}. We can see that as of the preliminary results, there is no clear pattern to determine which system is best. This suggests that the QA model, while trained on non-biomedical data, has learned at least as well as a domain-specific model to generalize matching questions to spans of text using the context of the match. Also, when looking at the accuracy of the models against the field of submissions, the non-ported NQ QA model is fairly strong - easily in the top third of submitted systems. This suggest that even general domain QA models can do a reasonable job on new domains, including hyper-specialized ones like biomedical literature.

\begin{table}[t]
    \caption{Domain portability for factoid biomedical QA}
    \label{tab:domain_port}
    \centering
     \begin{tabular*}{\textwidth}{@{\extracolsep{\fill}} c |ccc |ccc |ccc}
    \multicolumn{1}{c}{} & \multicolumn{3}{c}{No-Biomedical Fine-tuning} & \multicolumn{3}{c}{Biomedical Fine-tuning} & \multicolumn{3}{c}{}\\
    \multicolumn{1}{c}{} & \multicolumn{3}{c}{google-gold-input-nq} & \multicolumn{3}{c}{google-gold-input-ab} & \multicolumn{3}{c}{$\Delta$}\\    \multicolumn{1}{c}{} & Strict & Lenient & \multicolumn{1}{c}{MRR} & Strict & Lenient & \multicolumn{1}{c}{MRR} & Strict & Lenient &  \multicolumn{1}{c}{MRR} \\ \hline
    Batch 4 & 0.4706 & 0.6471 & 0.5255 & 0.4706 & 0.5882 & 0.5132 & - & 0.0589 & 0.0123\\
    Batch 5 & 0.2286 & 0.2857 & 0.2571 & 0.2857 & 0.3714 & 0.3057 & -0.0571 & -0.0857 & -0.0486 \\ \hline
    \end{tabular*}
    \vspace{0.1in}
\end{table}

Again, these results are preliminary, we can again look at previous BioASQ batches with more compete test annotations. Table~\ref{tab:prev_comp_domain_port} has the results. From here we can see that the biomedical specific model (google-gold-input) outperforms the domain general model (google-gold-input-nq) consistently, but not by a large margin. Furthermore, the domain general model is competitive with the previous state-of-the-art BioBERT models. These results present stronger empirical evidence that large-scale domain general models do port well to new domains.

\begin{table}[t]
    \caption{Performance on BioASQ Task 4 and 5b, Phase B averaged over all batches to measure domain portability. The bold system is the top scoring. $^*$This model is slightly different from the submitted system as it uses only gold snippets as input.}
    \label{tab:prev_comp_domain_port}
    \centering
     \begin{tabular*}{\textwidth}{@{\extracolsep{\fill}} c |ccc |ccc}
    \multicolumn{1}{c}{} & \multicolumn{3}{c}{BioASQ 4} & \multicolumn{3}{c}{BioASQ 5} \\
    \multicolumn{1}{c}{} & Strict & Lenient & \multicolumn{1}{c}{MRR} & Strict & Lenient & \multicolumn{1}{c}{MRR} \\ \hline
    BioBERT \cite{lee2019biobert} & \textbf{0.365} & 0.489 & \textbf{0.411} & 0.416 & 0.540 & 0.463 \\
    \hline
    google-gold-input &  0.311 & \textbf{0.540} & 0.400 & \textbf{0.458} & \textbf{0.615} & \textbf{0.520} \\
    google-gold-input-nq$^*$ &  0.302 & 0.488 & 0.376 & 0.451 & 0.603 & 0.509 \\\hline
    \end{tabular*}
    \vspace{0.1in}
    
\end{table}

It should be noted that we did not measure the effect of in-domain pre-training. BioBERT \cite{lee2019biobert} tested this and did find that for BioASQ 4-6 significant increases in factoid QA metrics could be achieved when using in-domain pre-training. This could suggest that pre-training and not fine-tuning are the keys to improving domain portability of BERT-based QA models.

\subsection{Error Propagation}

To test error propagation we used our main model: snippets as input; pre-trained BERT; fine-tuned on NQ; and further fine-tuned on BioASQ training data. We then tested two scenarios,
\begin{itemize}
    \item Gold inputs (google-gold-input): we used gold standard snippets generated by humans as input to the QA model. This is the standard setting for almost all participants in the track, as these were provided by the organizers.
    \item Noisy inputs (google-gold-pred): We used predicted snippets as input to the QA model. This was provided by \cite{pappas2019aueb}, a team that participated in 7b Phase A and whose document and snippet retrieval were the highest scoring submissions. Specifically, we used there BERT-based high-confidence document reranker plus snippet extractor.
\end{itemize}

Table~\ref{tab:err_prop} contains the results. We measure error propagation only for factoid QA for batches 3-5, which were the batches that we participated in. We can see from these results that feeding the QA model non-gold inputs leads to a dramatic drop in all metrics: from 7pts up to 14pts absolute. In one case (batch 5, strict accuracy), the metric is halved.

\begin{table}[t]
    \caption{Error propagation for factoid biomedical QA}
    \label{tab:err_prop}
    \centering
     \begin{tabular*}{\textwidth}{@{\extracolsep{\fill}} c |ccc |ccc |ccc}
    \multicolumn{1}{c}{} & \multicolumn{3}{c}{Gold Inputs} & \multicolumn{3}{c}{Noisy Inputs} & \multicolumn{3}{c}{}\\
    \multicolumn{1}{c}{} & \multicolumn{3}{c}{google-gold-input} & \multicolumn{3}{c}{google-gold-pred} & \multicolumn{3}{c}{$\Delta$}\\
    \multicolumn{1}{c}{} & Strict & Lenient & \multicolumn{1}{c}{MRR} & Strict & Lenient & \multicolumn{1}{c}{MRR} & Strict & Lenient &  \multicolumn{1}{c}{MRR} \\ \hline
    Batch 3 & 0.4138 & 0.6552 & 0.5023 & 0.3448	& 0.5517 & 0.4322 & 0.0690 & 0.1035 & 0.0701\\
    Batch 4 & 0.4706 & 0.7059 & 0.5495 & 0.3529 & 0.5882 & 0.4338 & 0.1177 & 0.1177 & 0.1157\\
    Batch 5 & 0.2857 & 0.3714 & 0.3167 & 0.1429	& 0.2857 & 0.2057 & 0.1428 & 0.0857 & 0.1110 \\ \hline
    \end{tabular*}
    \vspace{0.1in}
    
\end{table}

These results strongly suggest that when considering the QA system holistically -- retrieval followed by QA -- the largest bottleneck is the quality of the retrieval system, and not necessarily the QA model. For batch 3, our model was at the top or near the top for all metrics. However, for batches 4 and 5, our model was significantly lower than the top reporting system and we can see that error propagation is amplified for these batches. It would be useful to measure error propagation against the best reporting BioASQ models for these batches.

\section{Conclusion}

In this paper, we set out to investigate the domain portability of neural QA systems \cite{alberti2019bert} and to determine what is the impact of error propagation in end-to-end retrieval and QA systems. We found that even though our base QA model was trained on non-biomedical data, it was able to generalize matching questions to spans of text and gave very good results compared to systems that were trained with biomedical data. In addition, our results suggest that when using end-to-end QA systems the bottleneck is the quality of the retrieval system and not necessarily the QA model itself.

%
%
%
\bibliographystyle{splncs04}
\bibliography{bioasq}

\end{document}